\documentclass[letterpaper, 10 pt, conference]{ieeeconf}  

\IEEEoverridecommandlockouts                              
\usepackage{times}
\usepackage{graphicx}
\usepackage{amsmath,amssymb,amsopn,amstext,amsfonts}
\usepackage{cancel}
\usepackage[space]{cite}
\usepackage{pdfsync}
\usepackage{balance}
\usepackage{color}
\usepackage{mathtools}
\usepackage{algpseudocode}
\usepackage{algorithm}
\usepackage{bm}

\usepackage{diagbox}
\usepackage{float}
\usepackage{epstopdf}
\usepackage{pifont}
\usepackage{fixltx2e}
\usepackage{amsmath}
\usepackage{multirow}
\usepackage{url}
\usepackage{verbatim}
\usepackage{amssymb}
\makeatletter
\let\NAT@parse\undefined
\makeatother
\usepackage[linkcolor=black,citecolor=black,urlcolor=black,colorlinks=TRUE]{hyperref}
\usepackage{booktabs}
\usepackage{graphicx}
\usepackage{subcaption}
\usepackage{gensymb}
\usepackage{stfloats}
\usepackage{amsmath}
\usepackage{nccmath}

\graphicspath{{./FIGURES/}}
\DeclareGraphicsExtensions{.png,.jpg,.eps,.pdf}

\title{\LARGE \bf
	Reactive Aerobatic Flight via  Reinforcement Learning
}

\author{ Zhichao Han\textsuperscript{1,2},
	Xijie Huang\textsuperscript{2},
	Zhuxiu Xu\textsuperscript{2},
	Jiarui Zhang\textsuperscript{1,2},\\
	Yuze Wu\textsuperscript{1,2},
	Mingyang Wang\textsuperscript{1,2},
	Tianyue Wu\textsuperscript{1,2}
, and Fei Gao\textsuperscript{1,2}
	\thanks{
\emph{Corresponding author: Fei Gao}
	} 
	\thanks{\textsuperscript{1}State Key Laboratory of Industrial Control Technology, Zhejiang University, Hangzhou 310027, China.}
	\thanks{\textsuperscript{2}Huzhou Institute, Zhejiang University, Huzhou 313000, China.}	
	\thanks{E-mail:{\tt\small \{zhichaohan, fgaoaa\}@zju.edu.cn}}
}

\begin{document}

	\makeatletter
	\let\@oldmaketitle\@maketitle
	\renewcommand{\@maketitle}{\@oldmaketitle
		\begin{center}
			\includegraphics[width=1.0\linewidth]{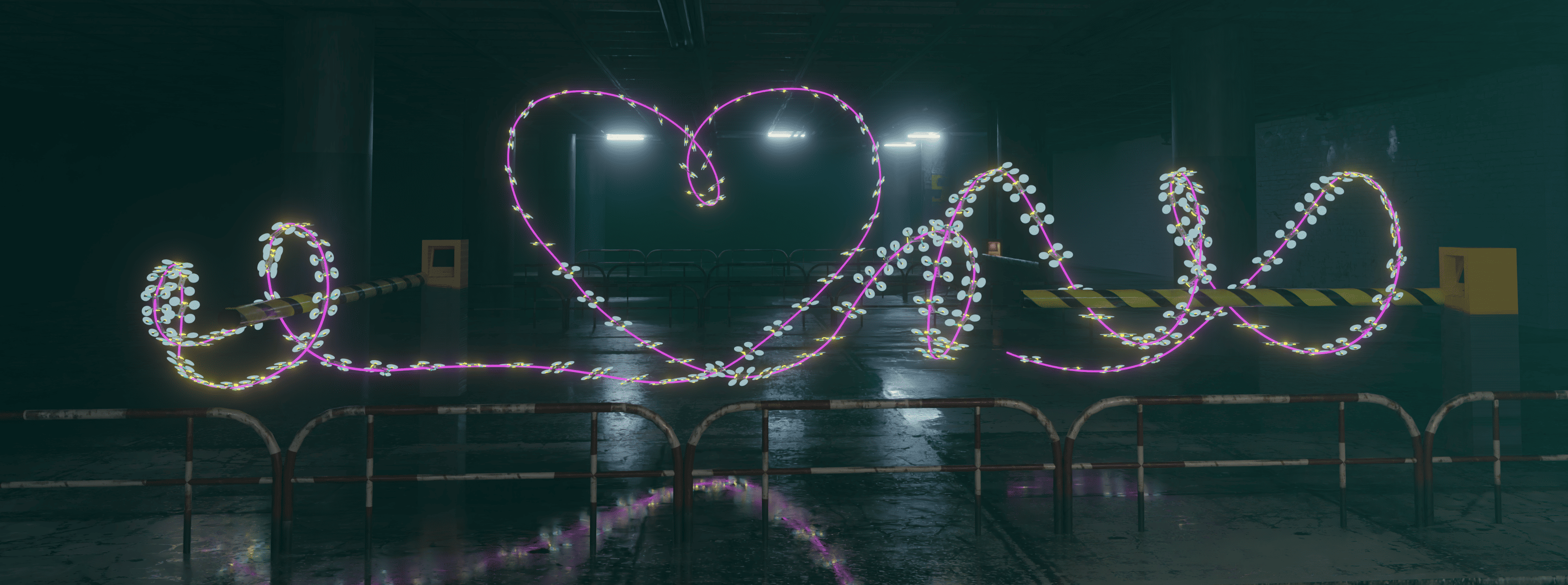}
		\end{center}
		\captionsetup{font={small}}
		\captionof{figure}{
			\label{fig:top}
			Visualization of the large-scale aerobatic flight trajectory, rendered using \textit{Blender}.
		} \label{fig:top}
	}
	\makeatother
	\maketitle
	\setcounter{figure}{1}
	\thispagestyle{empty}
	\pagestyle{empty}

\begin{abstract}
Quadrotors have demonstrated remarkable versatility, yet their full aerobatic potential remains largely untapped due to inherent underactuation and the complexity of aggressive maneuvers. Traditional approaches, separating trajectory optimization and tracking control, suffer from tracking inaccuracies, computational latency, and sensitivity to initial conditions, limiting their effectiveness in dynamic, high-agility scenarios.
Inspired by recent breakthroughs in data-driven methods, we propose a reinforcement learning-based framework that directly maps drone states and aerobatic intentions to control commands, eliminating modular separation to enable quadrotors to perform end-to-end policy optimization for extreme aerobatic maneuvers.
To ensure efficient and stable training, we introduce an automated curriculum learning strategy that dynamically adjusts aerobatic task difficulty.
Enabled by domain randomization for robust zero-shot sim-to-real transfer, our approach is validated in demanding real-world experiments, including the first demonstration of a drone autonomously performing continuous inverted flight while reactively navigating a moving gate, showcasing unprecedented agility.
\end{abstract}

\section{Introduction}
Unmanned Aerial Vehicles (UAVs), particularly quadrotors, have become increasingly integral across diverse sectors. While much research has focused on enhancing their stability and autonomy for conventional flight tasks, there is a growing interest in expanding their agility and pushing the limits of physical performance through aggressive maneuvers~\cite{wu2024whole,chen2020,Kaufmann}.  
Pushing these boundaries poses a range of fascinating and challenging research problems, such as enabling agile, visually impressive aerobatics reminiscent of avian flight (Fig. 1).
 These maneuvers demand rapid attitude shifts at high speeds, such as flips, a skill set traditionally exclusive to seasoned human pilots.
The control challenge is fundamentally rooted in the quadrotor’s inherent underactuation~\cite{mellinger2011minimum}, a characteristic that tightly couples its rotational dynamics with its translational movement.
Moreover, performing extreme maneuvers such as inverted flight or rapid flips at high speeds pushes the aircraft into highly unstable flight regimes, operating perilously close to its physical and hardware limitations. Within this demanding operational envelope, the system exhibits extreme sensitivity: even minute inaccuracies in control signals can precipitate catastrophic failure. 

Traditionally, researchers typically decompose aerobatic flight into two modules: trajectory optimization and tracking control~\cite{wmy,kaufmann2020deep}. In this framework, researchers first model the aerobatic trajectory optimization as a nonlinear optimal control problem based on the UAV’s kinematic model, incorporating dynamic constraints to ensure physical feasibility of the trajectory. Subsequently, they implement a feedback controller to execute trajectory following and complete the aerobatic maneuver.
However, such methods have disadvantages in the following three aspects:
1. Tracking error. While the separation of planning and control modules seems intuitive, trajectory optimization typically relies on idealized kinematic models. Real-world  noises  prevent trajectories from being followed with perfect precision~\cite{kamel2017model}, particularly during large-attitude maneuvers, potentially leading to control divergence.
2. Latency. Beyond the inherent delays introduced by modular separation, full-state trajectory optimization in $\mathbb{SE}(3)$ is computationally intensive, making efficient online replanning challenging~\cite{hanse3}. This computational burden severely restricts the system’s reactive capabilities—a crucial disadvantage in highly challenging dynamic aerobatic scenarios, such as inverted flight through moving gates. 
3. Optimality.
The quality of solutions from traditional nonlinear trajectory optimization methods heavily depends on the initial values, with poor initial values often leading to unsatisfactory local minima~\cite{8967980}.

In recent years, data-driven methods~\cite{loquercio2021learning,song2023reaching,kaufmann2023champion} have revolutionized robotics research, offering elegant end-to-end policy optimization without requiring explicit modular decomposition. Inspired by these advances, we explore the use of  online deep reinforcement learning to push the boundaries of UAV capabilities, enabling drones to achieve extreme aerobatic maneuvers.
We train aerobatic policies in simulations where control commands are generated directly from aerobatic intent and drone state inputs, eliminating the traditional separation between trajectory optimization and tracking control modules. 
Unlike conventional model-based optimization approaches, our model-free policy search learns directly from vast datasets collected through continuous environmental interaction, effectively avoiding shallow local optima while elegantly handling complex uncertainties that resist explicit modeling.
Furthermore, lightweight neural networks enable real-time  control with instantaneous reactions to dynamic changes in aerobatic waypoints, which is challenging for computationally intensive trajectory optimization.
However, the precise control required for aerobatics makes learning directly from a fixed initial state challenging, hindering training efficiency and potentially compromising the quality of the converged solution.  
Therefore, based on curriculum learning~\cite{graves2017automated,florensa2017reverse,portelas2020automatic}, we innovatively introduce a progressive learning strategy that dynamically selects appropriate reset states for the quadrotor based on the network’s current proficiency, thereby intelligently modulating the difficulty of aerobatic tasks to ensure stable convergence.
   Furthermore, to bridge the sim-to-real gap, we employ domain randomization on the idealized dynamics model, enhancing robustness to unforeseen uncertainties and enabling zero-shot sim-to-real transfer.  Real-world experiments demonstrate our autonomous drone successfully performing large-scale aerobatic maneuvers, including reactive adaptation to dynamic environments, showcased by rapidly and continuously traversing a moving gate while maintaining an inverted flight attitude.
Overall, the main contributions of this paper can be
summarized as follows:
\begin{itemize}	
	\item [1)]
	We present a reinforcement learning-based policy approach for achieveing the extreme motor skills of large-scale aerobatic flight.
	\item [2)]
	We introduce automated curriculum learning to adaptively adjust the difficulty of aerobatic training, ensuring efficient and stable network convergence.
	\item [3)] 
	We achieve zero-shot sim-to-real transfer and conduct real-world experiments to validate the effectiveness of our system.
\end{itemize}	

\section{Related Work}
In traditional approaches, researchers typically first plan a dynamically feasible trajectory and then employ a standard controller, such as model predictive control (MPC) or a PID-based method, to achieve trajectory tracking. Chen et al.~\cite{chen2020} simplify aerobatic trajectory planning through manually designed rules, successfully achieving multiple flip maneuvers. However, this approach introduces numerous rule-based parameters requiring extensive manual tuning for motion shape and velocity, significantly limiting the optimality.
Romero et al.~\cite{romero2022model} model the optimal control problem by discretizing the motion process to generate highly agile, high-speed trajectories. However, this approach faces a significant trade-off between solution quality and computational efficiency, as higher discretization accuracy leads to dramatically increased computational complexity.
Although some methods~\cite{hanse3,lu2024trajectory} leverage differential flatness to simplify the problem and improve efficiency, trajectory optimization on $\mathbb{SE}(3)$ remains computationally expensive, hindering real-time and efficient replanning. Moreover, flatness-based mappings exhibit singularities at large attitude angles~\cite{morrell2018differential}, posing additional challenges for generating dynamically feasible aerobatic trajectories.

Recently, learning-based approaches have also attracted increasing attention in quadrotor motion planning and control. Loquercio et al.~\cite{loquercio2021learning} use neural networks to learn high-speed obstacle avoidance trajectories from teacher demonstrations, but this method requires extensive expert data that is difficult to obtain and still relies on a separate trajectory tracker. Zhang et al.~\cite{zhang2024back} develop an end-to-end navigation system based on differentiable physics, yet the generated trajectories exhibit relatively low accelerations and lack sufficient aggressiveness. Other studies~\cite{song2023reaching,kaufmann2023champion} explore reinforcement learning for autonomous drone racing and achieve performance comparable to professional human pilots, but these methods do not address aerobatic maneuvers involving rapid attitude changes such as continuous flips.
Kaufmann et al.~\cite{Kaufmann} employ imitation learning to achieve aerobatic flight control, but this approach heavily depends on the quality of expert demonstrations and still requires high-quality reference trajectories computed offline, making it difficult to handle the highly challenging dynamic scenarios considered in this work.
\section{Methodology}
\subsection{Quadrotor Dynamics}
\label{sec:dynamics}
The state of an underactuated UAV is described by 
$\bm{s} = [\bm{p}, \bm{q}, \bm{v}, \bm{\omega}]$, and its ideal kinematic equations are given by:
\begin{align}
\dot{\bm{p}} &= \bm{v}, \dot{\bm{v}} = \bm{q}\otimes\bm{z}_{B} + \bm{g},\nonumber\\
\dot{\bm{q}} &=\frac{1}{2}[\bm{\omega}]_\times \cdot \bm{q}, \dot{\bm{\omega}} = \mathbf{J}^{-1}(\bm{\tau}-\bm{\omega}\times\mathbf{J}\bm{\omega}). \label{eq:dynamics}
\end{align}
Here, 
$\bm{p}$, $\bm{v}$ 
and $\bm{q} = [q_w, q_x, q_y, q_z]$
 denote the position, velocity, and unit quaternion of the quadrotor expressed in the world frame, respectively.
 The notation  $[\cdot]_\times$
 indicates the skew-symmetric matrix form of a vector.
$\bm{g} = [0,0,-9.81]^{\rm T}$ is the constant gravitational acceleration. $\bm{\omega}$
 denotes the angular velocity in three axes and  $\mathbf{J}$
 is the inertia matrix.
   $\bm{z}_B=[0,0,T_r]^{\rm T}$ 
 represents the mass-normalized thrust in the robot body frame and 
$\bm{\tau}$
 is the three-dimensional torque. 
 Moreover, once the single mass-normalized rotor thrusts $[f_1,f_2,f_3,f_4]$ are obtained, the collective thrust $T_r$ and $\bm{\tau}$ can be directly and easily computed~\cite{song2023reaching}.
\subsection{Problem Formulation}

We  formulate the large-scale aerobatic flight problem within our reinforcement learning framework as a general infinite-horizon Markov Decision Process (MDP). An MDP is defined by a tuple \((\mathcal{S}, \mathcal{A}, \mathcal{P}, r, \gamma)\), where \(\mathcal{S}\) denotes the state space, \(\mathcal{A}\) the action space, \(\mathcal{P}: \mathcal{S}\times\mathcal{A}\times\mathcal{S}\rightarrow[0,1]\) the state transition probability, \(r:\mathcal{S}\times\mathcal{A}\rightarrow\mathbb{R}\) the reward function, and \(\gamma\in[0,1)\) the discount factor. The objective of reinforcement learning is to find an optimal policy \(\pi^*\) that maximizes the expected cumulative discounted reward:
\begin{align}
\pi^* = \arg\max_{\pi}\mathbb{E}_{\tau\sim\pi}\left[\sum_{t=0}^{\infty}\gamma^t r(\bm{s}_t,\bm{u}_t)\right],
\end{align}
where \(\tau=(\bm{s}_0,\bm{u}_0,\bm{s}_1,\bm{u}_1,\dots)\) denotes a trajectory generated by policy \(\pi\)
and $\bm{u}$ represents the action output at each step.
Next, we instantiate the  observation, the action space and the reward functions.
\subsubsection{Observation Space}
\label{sec:obs}
Similar to the formulation presented in our previous work~\cite{wmy}, the large-scale aerobatic flight tasks considered in this paper are modeled as controlling a UAV to start from an initial state $\bm{s}_0$ sampled from a set of predefined starting conditions $\mathcal{S}_{start}$, and then swiftly traverse a sequence of moving $\mathbb{SE}(3)$ aerobatic waypoints $\mathbf{Q}_{ws}$ encoding human-designed flight intentions at desired tilt angles.
Under this problem definition, the observation input to our policy network is structured as a composite vector \( \bm{O} = [\bm{O}_{env}, \bm{O}_{ego}] \), integrating both environmental observations $\bm{O}_{env}$ that encapsulate the intended aerobatic maneuvers and the robot's proprioceptive state $\bm{O}_{ego}$ information. Specifically, without loss of generality, the environmental observation \( \bm{O}_{env}=[\delta \widetilde{\bm{p}}_1, \widetilde{\bm{q}}_1, \delta \widetilde{\bm{p}}_2, \widetilde{\bm{q}}_2]\) is formulated as a concatenation of the next immediate aerobatic waypoint $[\delta \widetilde{\bm{p}}_1, \widetilde{\bm{q}}_1]$ and the subsequent waypoint $[\delta \widetilde{\bm{p}}_2, \widetilde{\bm{q}}_2]$ thereafter.
This representation dynamically updates as the UAV progresses along the predefined aerobatic track, ensuring that the policy continuously receives relevant and timely guidance for upcoming maneuvers.
Here,  $\delta \widetilde{\bm{p}}$ represents the three-dimensional coordinates of the aerobatic point in the current body coordinate system of the quadrotor, while $\widetilde{\bm{q}}$ represents the attitude of the aerobatic point, expressed as a unit quaternion. 
The robot's proprioceptive observation is defined as a composite vector \( \bm{O}_{ego} = [\bm{p},\bm{q},\bm{v}_B, \bm{u}_{last}] \) where
$\bm{v}_B$  is the velocity expressed in the robot's body frame.
Moreover, \(\bm{u}_{last}\) denotes the action output from the previous network inference, serving as a reference to ensure that the current network output does not deviate drastically from the preceding action. 
\subsubsection{Action Space}
Our policy network is designed to perform inference continuously at a fixed frequency of 100 Hz, outputting a four-dimensional action vector $\bm{u} = [T_r, \omega_x, \omega_y, \omega_z]$ consisting of mass-normalized collective thrust and body rates along three axes. 
These action commands are directly fed into the onboard flight controller integrated on the hardware platform, which subsequently converts them into motor speeds.
We select this particular action space for two primary reasons. First, it closely mimics human pilot behavior, as professional drone pilots typically control quadrotors using collective thrust and body rate commands via remote controllers. Second, this choice achieves a balance between agile maneuverability and manageable sim-to-real transferability.
Higher-level control inputs, such as position or velocity commands, often sacrifice agility and responsiveness, whereas lower-level inputs, such as direct motor speeds, tend to introduce significant discrepancies between simulation and real-world performance~\cite{song2023reaching}.
\subsubsection{Reward Function}
Our reward function is defined as a weighted sum of individual sub-rewards:
\begin{align}
r = w_0r_{aer} + w_1r_{act}+w_2r_{act\_change} + w_3r_{yaw} \label{eq:origin_reward}
\end{align}Below, we provide a detailed discussion of the specific forms of these sub-rewards.
The reward \( r_{aer} \) is used to evaluate   aerobatic maneuvers. Before introducing this reward in detail, we first briefly define the criterion for determining the completion of an aerobatic maneuver. Given the current aerobatic waypoint $[\widetilde{\bm{p}}, \widetilde{\bm{q}}]$ to be passed, the maneuver is considered  completed
if the UAV crosses the local Y-O-Z plane of the waypoint along the X-axis direction defined by the waypoint's orientation, and if the UAV's distances from the waypoint along both the Y-axis and Z-axis directions do not exceed a predefined threshold \( L \).
Then, $r_{aer}$ is defined as follows:
\begin{align}
&\bm{p}_{A} =[p_{x,A},p_{y,A},p_{z,A} ]={\widetilde{\bm{q}}}^{-1}\otimes(\bm{p}_W-\widetilde{\bm{p}}) \nonumber\\
&p_{error} = ||\bm{p}_{A}||, \theta_{error} = \arccos( 
(\widetilde{\bm{q}}\otimes\bm{e}_3)^{\rm T}
(\bm{q}_W\otimes\bm{e}_3) \nonumber\\
&r_{aer} = 
(\mathcal{F}(p_{error}) + 
\mathcal{F}(
\theta_{error})+c)\cdot \\
&\mathcal{C}(p_{x,A}>0
\ \text{and} \ p_{x,A}^{last} \leq 0 
\ \text{and} \ 
|p_{y,A}|\leq L 
\ \text{and} \ |p_{z,A}|\leq L)
 \nonumber \label{eq:orgin_aer}
\end{align}where \(\mathcal{C}(\cdot)\) is an indicator function that returns 1 when the aerobatic maneuver is completed and 0 otherwise. $c$ is a constant reward score.
The first term scores the aerobatic maneuver, encouraging the flight to match the desired aerobatic position and inclination.
The activation function $\mathcal{F}(x)$ is designed as a concatenation of linear and exponential functions, thereby magnifying the reward gradient near the optimal state and encouraging the policy to converge more precisely toward the desired aerobatic configuration:
\begin{small}
\begin{align}
\mathcal{F}(x) = (a-x)\mathcal{C}(x>a-b)+(b - 1 + e^{\,a - x - b})\mathcal{C}(x\leq a-b),
\end{align}
\end{small}where \(a\) and \(b\) are predefined constants.
The terms \( r_{act} \) and \( r_{act\_change} \) are metrics designed to quantify motion smoothness by penalizing the magnitude of actions and the changes between consecutive actions, respectively:
\begin{align}
r_{act} = \sum_{\mu = \{x,y,z\}}|\omega_\mu|,r_{act\_change} = ||\bm{u} - \bm{u}_{last}||_2.
\end{align}
The term \( r_{yaw} \) denotes the yaw alignment reward, which encourages the UAV's heading direction to align closely with its velocity vector. This alignment results in more natural flight trajectories that better match human pilots' intuitive control strategies and facilitates first-person-view recording:
\begin{align}
 r_{yaw} = -\arccos(\frac{\bm{v}_B - (\bm{v}_B\cdot \bm{e}_3) \bm{e}_3}{||\bm{v}_B - (\bm{v}_B\cdot \bm{e}_3) \bm{e}_3||}\cdot \bm{e}_1),
\end{align}
where $\bm{e}_3 = [0,0,1]$ and $\bm{e}_1 = [1,0,0]$.
 Physically, this reward penalizes the angle between the UAV's velocity projection onto its body-fixed X-O-Y plane and the body X-axis.

\subsubsection{Termination Condition}
The agent will terminate when the simulation step exceeds a predefined threshold or when the agent moves outside a local bounding box defined with respect to the coordinate frame of the current aerobatic waypoint.
\subsection{Policy Optimization}
\begin{figure*}[t]
	\centering
	\includegraphics[width=1\linewidth]{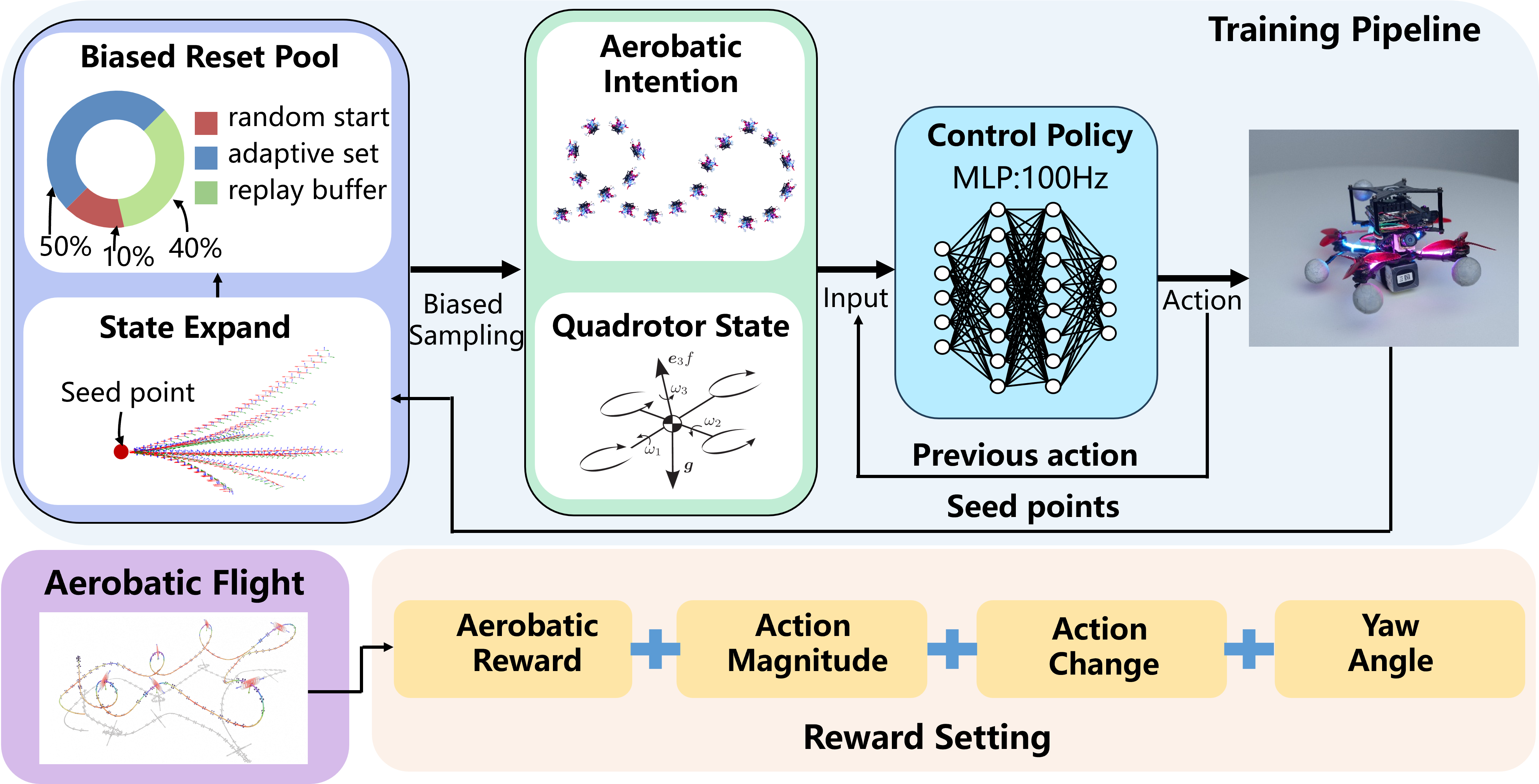}
	\captionsetup{font={small}}
	\caption{The training framework.}
	\vspace{-0.5cm}
	\label{fig:pipeline}
\end{figure*}
Our training framework is shown in Fig.~\ref{fig:pipeline}.
The policy network is an MLP that encodes observations into a 512-dimensional latent vector, followed by four fully connected layers of sizes 512, 512, 256, and 128. A final projection layer maps the features to the normalized action space using a Tanh activation, constraining outputs to \([-1, 1]\). These actions are then rescaled according to predefined mean and amplitude parameters to produce the actual control commands for the robot. We train the policy using Proximal Policy Optimization (PPO)~\cite{ppo}, a policy-gradient method that constrains policy updates for stable and efficient learning. 

Under our problem formulation, the most critical reward \( r_{aer} \) is sparse, as it can only be obtained upon successful completion of an aerobatic maneuver. 
Although the aerobatic flight task requires the UAV to start from an initial set $\mathcal{S}_{start}$ and sequentially pass through a series of aerobatic waypoints, directly training under this setting is highly inefficient. This inefficiency arises primarily due to the sparsity of the reward and the stringent control accuracy required for highly agile aerobatic maneuvers, which typically necessitate an impractically large amount of data to sufficiently explore feasible control commands, severely impairing the convergence efficiency of the model. To address this issue, 
based on automatic curriculum learning~\cite{graves2017automated,florensa2017reverse,portelas2020automatic},
we design a progressive training strategy to improve training efficiency without sacrificing optimality.

\begin{algorithm}[tbp]
	\caption{Biased Reset-Based Training Strategy}
	\label{alg:acl}
	\begin{algorithmic}[1] 
		\State \textbf{Definition:} 
		aerobatic waypoints $\mathbf{Q}_{ws}$,  critic network $V$,   expansion steps $K$,		simulation environment $\mathcal{E}$
		\State $\mathcal{S}_{buffer} \gets \emptyset$
		\State $\mathbf{x}^g \gets$\texttt{Extract\_Goals}($\mathbf{Q}_{ws}$)
		\State $\mathbf{x}^g_{flat} \gets$ \texttt{State\_to\_Flat}($\mathbf{x}^g$)
		\State  
		$\mathbf{x}^{g,exp}_{flat} \gets$ \texttt{Expand\_Flats}($\mathbf{x}^g_{flat}$, $K$, $\Delta t$) \Comment{Eq. (\ref{eq:reverse})}
		\State
		$\mathcal{S}_{curr}\gets$
		\texttt{Flat\_to\_State}($\mathbf{x}^{g,exp}_{flat}$)
		\State $\mathcal{S}_{buffer} \gets \mathcal{S}_{buffer} \cup \mathcal{S}_{curr}$
		\For{each episode}
		\State Sample  $\rho \sim \text{Uniform}(0,1)$\Comment{Biased sampling.}
		\If{$\rho < \rho_1$}
		\State Sample initial state $\bm{s}_0$ from $\mathcal{S}_{curr}$
		\ElsIf{$\rho < \rho_1 + \rho_2$}
		\State Sample initial state $\bm{s}_0$ from $\mathcal{S}_{buffer}$
		\Else
		\State Sample initial state $\bm{s}_0$ from $\mathcal{S}_{start}$
		\EndIf
		\State $\mathcal{E}$\texttt{.reset}($\bm{s}_0$)
		\State $\pi,V,states$ $\gets$ $\mathcal{E}$\texttt{.train}($\pi$,$V$)
		\State $\mathbf{x}^m \gets V$.\texttt{Evaluate\_and\_Select}($states$)
		\State $\mathbf{x}^m_{flat} \gets$ \texttt{State\_to\_Flat}($\mathbf{x}^m$)
		\State  		$\mathbf{x}^{m,exp}_{flat} \gets$ \texttt{Expand\_Flats}($\mathbf{x}^m_{flat}$, $K$, $\Delta t$) \Comment{Eq. (\ref{eq:reverse})}
		\State
		$\mathcal{S}_{curr}\gets$
		\texttt{Flat\_to\_State}($\mathbf{x}^{m,exp}_{flat}$)
		\State $\mathcal{S}_{buffer} \gets \mathcal{S}_{buffer} \cup \mathcal{S}_{curr}$
		\EndFor
	\end{algorithmic}
\end{algorithm}

Intuitively, states closer to aerobatic waypoints are typically easier for the UAV to achieve desired maneuvers. Then, rather than uniformly sampling initial states, we dynamically select reset states based on their relative difficulty, guiding the agent from simpler scenarios toward increasingly challenging ones.  The pseudo-code of the algorithm is shown in Alg.~\ref{alg:acl}.
Similar to the observation space definition in 
Sect.~\ref{sec:obs}, we define a goal state \( \mathbf{x}^g = [\bm{p}^g, \bm{q}^g, \bm{v}^g] \), where \( \bm{p}^g \) and \( \bm{q}^g \) are equal to the desired aerobatic waypoint's position and orientation, respectively, and \( \bm{v}^g \) is a velocity vector sampled within a predefined range.
Since the distance between two states cannot accurately measure the true cost of the transition between them, we do not directly perform random sampling in the state space. Instead, we estimate the previous state by randomly generating control inputs. Furthermore, to avoid the inversion of the nonlinear kinematic Eq. (\ref{eq:dynamics}), we apply differential flatness to convert it into a linear model, thereby simplifying the calculation.
 Specifically, we define the flat state as:
$\mathbf{x}^{flat} = [\bm{p}, \bm{v}, \bm{a}]$
where \( \bm{a} \) is the acceleration.  Given the flat state at time \( t \) and a randomly sampled jerk input \( \bm{j} \), we analytically compute the flat state at the previous timestep \( t-1 \):
\begin{align}
	\bm{a}_{t-1} &= \bm{a}_{t}-\bm{j}_{t-1}\Delta t,\nonumber\\
	\bm{v}_{t-1}&=\bm{v}_{t} - \bm{a}_{t-1}\Delta t-\frac{1}{2}\bm{j}_{t-1}\Delta^2 t, \nonumber\\
	\bm{p}_{t-1}&=\bm{p}_{t}-\bm{v}_{t-1}\Delta t-
	\frac{1}{2}\bm{a}_{t-1}\Delta^2t-\frac{1}{6}\bm{j}_{t-1}\Delta^3t, \label{eq:reverse}
\end{align}
where $\Delta t = 0.01s$ is the expandsion time interval.
Assuming the UAV's heading aligns with the projection of its velocity onto the body-fixed X-O-Y plane, the robot rotation can be directly derived from the flat state:
\begin{align}
\bm{z}^{b} = \frac{\bm{a} - \bm{g}}{||\bm{a} - \bm{g}||},
\bm{x}^{b}=
\frac{\bm{v} - (\bm{v}^{\rm T}\bm{z}^{b}) \bm{z}^b}{||\bm{v} - (\bm{v}^{\rm T}\bm{z}^{b}) \bm{z}^b||},
 \bm{y}^b = \bm{z}^b \times \bm{x}^b.
\end{align}Here, \( \bm{x}^b \), \( \bm{y}^b \), and \( \bm{z}^b \) are the X, Y, and Z axes of the body-fixed coordinate system, which can be utilized to derive the quaternion.
By recursively applying this backward integration procedure Eq. (\ref{eq:reverse}), we expand the state and obtain a diverse set of candidate reset states at varying temporal distances from the goal state.
To dynamically adjust reset state difficulty,
at each training iteration,
 we use the critic network to evaluate states and expand those states $\mathbf{x}^m$ of moderate difficulty (seed points) to form the adaptive reset set \(\mathcal{S}_{curr}\) for the next iteration.
 Additionally, to mitigate catastrophic forgetting, we maintain an experience replay buffer \(\mathcal{S}_{buffer}\) containing previously recorded states.
In the implementation, we employ a biased sampling strategy for state resets: with a $\rho_1$ probability, samples are drawn from \(\mathcal{S}_{curr}\), a $\rho_2$ probability from \(\mathcal{S}_{buffer}\), and the remaining probability from $\mathcal{S}_{start}$.

\subsection{Sim-to-Real Transfer}
In real-world scenarios, numerous factors such as time delays and uncertain noise are typically neglected by idealized kinematic models, resulting in discrepancies between simulation and reality. To mitigate this sim-to-real gap, we employ domain randomization in the simulator to ensure that the distribution of training data encompasses real-world conditions, thereby enhancing the generalization capability of the learned policy. Specifically, we first refine the original idealized model by incorporating parasitic air drag effects~\cite{drag}:
\begin{align}
\dot{\bm{v}} = \bm{q}\otimes\bm{z}_B + \bm{g} - \bm{q}\otimes\mathbf{D}\otimes\bm{q}^{-1}||\bm{v}||\bm{v},
\end{align}
where \( \mathbf{D} = \text{diag}(d_x, d_y, d_z) \) is a constant diagonal matrix composed of rotor-drag coefficients, which can be approximately estimated from  parameters such as cross-sectional area and air density.
To enhance environmental diversity and improve the robustness of the learned policy against potential inaccuracies in the aerodynamic drag model, we apply randomization to the drag coefficient matrix \( \mathbf{D} \) during training. Specifically, each element of \(\mathbf{D} \) is perturbed randomly within ±50\% of its nominal value. Similarly, to increase the policy's robustness to uncertainties in the system's response to control actions (mass-normalized collective thrust and body rates), we introduce random perturbations to the control inputs before performing forward simulation of the equations of motion, with variations uniformly sampled within ±20\% of their original values.
To deal with delays, we do not use the neural network’s instantaneous output as input to the kinematic model; instead, we average the outputs over a past interval (\(\delta_{interval} = [-t_0 - t_L, -t_0]\)), where \(t_L = 80\,\mathrm{ms}\) is fixed, and \(t_0\) is an offset from the current time. During training, \(t_0\) is uniformly sampled from \([4\,\mathrm{ms}, 36\,\mathrm{ms}]\) to enhance policy robustness and generalization to varying latency conditions.
\section{Evaluations}
In this section, 
we first conduct an ablation study to demonstrate the effectiveness of our automatic curriculum learning strategy in improving learning efficiency and convergence. Next, we compare our method with a recent aerobatic trajectory-planning baseline to highlight its superior optimality and responsiveness. We also analyze the impact of 
varying speeds of dynamic aerobatic waypoint on performance.
 Finally, we validate our approach through challenging real-world aerobatic flights. Besides, we use the open-source Flightmare simulator~\cite{flightmare} for training.
Simulated experiments run on an Intel Core i7-10700 CPU and NVIDIA RTX 2060 GPU, while real-world inference uses a Jetson Orin NX.
\subsection{Simulations}
\subsubsection{Ablation Studies}
\begin{figure}[t]
	\centering
	\includegraphics[width=1\linewidth]{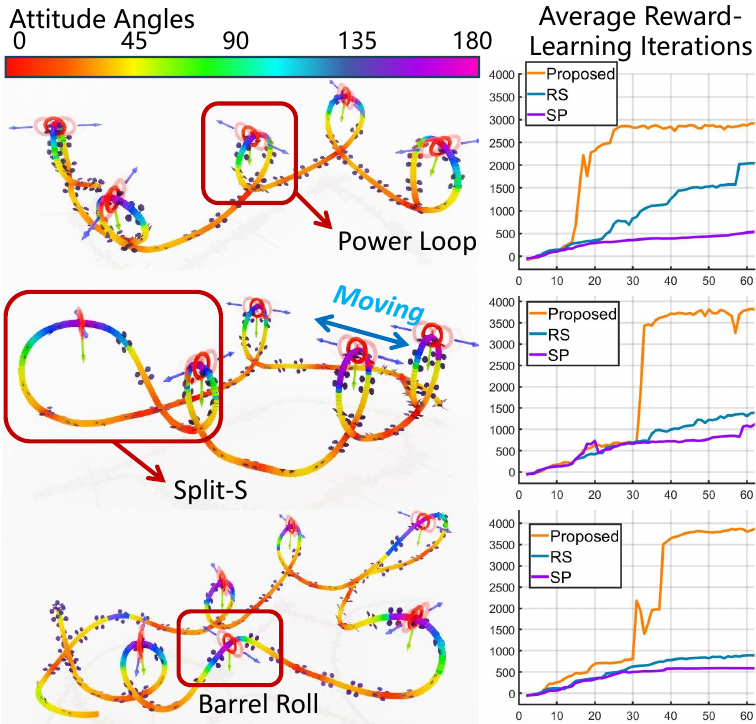}
	\captionsetup{font={small}}
	\caption{
	Trajectory visualizations from ablation studies and reward evolution curves during  training processes. The red circles indicate the positions of the moving aerobatic point, with the Z-axis consistently oriented downward. The attitude angle refers to the angle between the UAV’s Z-axis and the vertically upward axis.
	}
	\vspace{-0.5cm}
	\label{fig:abla1}
\end{figure}
In this experiment, we design three aerobatic race tracks as illustrated in Fig.~\ref{fig:abla1}, incorporating a series of challenging aerobatic maneuvers including Power Loop, Barrel Roll, and Split-S~\cite{kaufmann2020deep,wmy}.  
The aerobatic flight task is defined as controlling the UAV rapidly through a sequence of moving aerobatic waypoints from a randomly initialized state \(\bm{s}_0 \in \mathcal{S}_{start}\). Specifically, during experiments, each aerobatic waypoint oscillates along its local Y-axis with a maximum displacement of 1.5 m and a velocity uniformly sampled between 0 and 1.0 m/s.
Here, we compare three training strategies:
1) \textbf{Sparse Reward (SP)}: 
directly address this flight mission by reseting agents within $\mathcal{S}_{start}$ during the training  and  using the original reward Eq. (\ref{eq:origin_reward}).
2) \textbf{Reward Shaping (RS)}: building upon SP, the original sparse  aerobatic reward Eq.(\ref{eq:orgin_aer}) is augmented with heuristic dense rewards $r_{rs}$ to guide training:
\begin{align}
r_{rs} = w_{sp}(p_{error}^{last}-p_{error}) + w_{sq}(\theta_{error}^{last}-\theta_{error}).
\end{align}Its physical meaning is the previous position and attitude error minus the current error, where $w_{sp}$ and $w_{sq}$ are weights.
3) \textbf{Proposed}: agents are trained using our progressive learning scheme, eliminating the need for manually designed reward shaping.
Throughout the training, we continuously evaluate the policy by deploying the UAV from  \(\bm{s}_0 \in \mathcal{S}_{start}\) and record the averaged original reward Eq. (\ref{eq:origin_reward}) for quantitative comparison, as presented in Fig.~\ref{fig:abla1}.
The results illustrate that our method achiev{\tiny }es the fastest training speed and best convergence performance across all evaluated trajectories. This advantage primarily stems from the  progressive training strategy, which dynamically adjusts the reset states during training and implicitly modulates the difficulty of the aerobatic maneuvers. Consequently, even in early training stages, the agent obtains sufficient opportunities to receive meaningful aerobatic rewards, effectively mitigating the exploration difficulties arising from sparse reward settings.
In contrast, the \textbf{SP} baseline suffers from severely inefficient exploration, as the sparse reward structure makes it difficult for the network to adequately identify valuable feedback signals. 
Although the \textbf{RS} baseline  provides heuristic guidance through reward densification, its lack of a progressive task difficulty curriculum results in lower training efficiency on challenging acrobatic control problems compared to our approach.
Moreover, \textbf{RS} compromises optimality, as the explicit, manually designed heuristic guidance often induces overly greedy policy learning, making convergence to undesired 
suboptimal solution more likely.  
In contrast, our approach preserves the original reward structure of the problem and thus does not sacrifice optimality.

\subsubsection{Benchmarks}
In this section, we compare our proposed approach against a recent state-of-the-art method~\cite{wmy} in aerobatic motion planning that has demonstrated the potential to achieve human-level aerobatic flight performance. This  method formulates the aerobatic trajectory generation problem as a nonlinear spatio-temporal optimization in flat-output space, employing a compact polynomial-based trajectory representation to reduce problem dimensionality. 
Subsequently, a separate trajectory-tracking controller is used to follow the optimized trajectories. However, the computational cost of the trajectory optimization module remains prohibitively expensive, typically requiring hundreds of milliseconds, and sometimes even exceeding one second, leading to substantial delays incompatible with online replanning. Consequently, this method cannot be directly applied to scenarios with dynamically changing aerobatic waypoints that demand rapid, real-time responses, as considered in our study.
In addition to latency, this method
relies on many manually tuned parameters  which  are not  universally applicable across different planning problems,  limiting the scalability  to scenarios requiring continuous replanning.
Therefore, to facilitate a fair comparison, we adopt an experimental setup similar to  the  work~\cite{wmy}, specifically considering scenarios with only static  aerobatic waypoints, thereby eliminating the necessity of online replanning during execution.
\begin{figure}[t]
	\centering
	\includegraphics[width=1.0\linewidth]{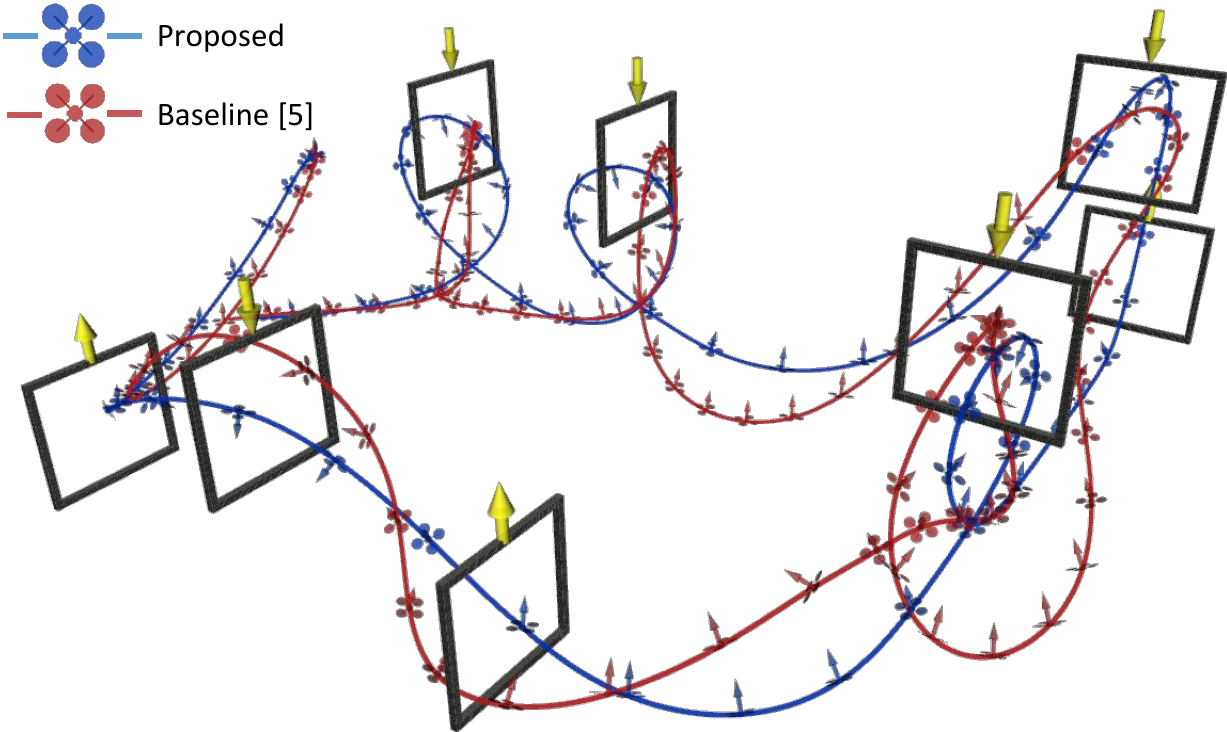}
	\captionsetup{font={small}}
	\caption{
		Visualization of flight trajectories. The center of each black gate denotes the position of the aerobatic point, while the yellow arrow on top indicates its Z-axis direction.
	}
	\vspace{-0.5cm}
	\label{fig:benchmark}
\end{figure} 
The flight trajectories of both methods are presented in Fig.~\ref{fig:benchmark}.
Moreover, we randomly select 20 sets of initial points to perform a quantitative comparative analysis. We record the average flight time (F.T), average position (Aer.P) and attitude (Aer.A) errors at aerobatic points, as well as average trajectory tracking errors in position (TK.P) and attitude (TK.A), as shown in Table.~\ref{tab:benchmarks}. Following the definition presented in the work~\cite{wmy}, the attitude error is measured as the angle between the actual and desired UAV body-frame Z axes.
This result highlights substantial advantages of our proposed approach in achieving superior optimality. Specifically, our method completes aerobatic maneuvers more rapidly while also maintaining lower position and attitude errors at critical aerobatic points. Such improved accuracy indicates that the UAV controlled by our approach is better aligned with the  aerobatic intention.
The baseline~\cite{wmy} exhibits inferior performance primarily due to two fundamental limitations. First, its trajectory planning is formulated as a nonlinear optimization problem, making it prone to suboptimal local minima. Additionally, the algorithm represents trajectories using piecewise polynomials, a representation lightweight yet inherently restrictive in terms of degrees of freedom. Consequently, it struggles to fully exploit the UAV’s dynamic and maneuvering capabilities. 
Although increasing the number of segments may help mitigate this issue, it also substantially increases the dimensionality of the optimization problem, thereby rendering it considerably more challenging to solve.
Second, the baseline~\cite{wmy} adopts a hierarchical framework with separate trajectory planning and tracking stages.  Despite having a robust trajectory-tracking controller,  tracking errors still persist, especially prominent during aggressive aerobatics.
For example, our experiments indicate that the baseline can produce maximum tracking errors up to 0.95m in position and 72.1° in attitude  during maneuvers.
In contrast, our approach completely bypasses trajectory planning and tracking layers by directly predicting  control commands from the learned policy network, fundamentally eliminating this issue. Finally, regarding computational efficiency, the baseline method requires approximately 1.5 seconds for trajectory planning, which severely limits its applicability in real-time dynamic scenarios. In contrast, our proposed method exhibits significantly higher efficiency, with a response time of only 0.3 ms.

Naturally, we are also interested in evaluating how our approach performs when the aerobatic point moves dynamically at high speeds. During training, the aerobatic point velocity is randomly set between 0 and 2 m/s. To thoroughly assess our policy's robustness under varying dynamic conditions, we conduct 50 test flights with different aerobatic point velocities, summarized in Table~\ref{tab:dynamic}. Results reveal that although performance gradually declines as the aerobatic point velocity increases, our approach remains remarkably robust. For instance, 
even under an extremely dynamic scenario with a velocity of 3.4 m/s, which also exceeds the maximum speed encountered during training, our policy still achieves a 78\% success rate and an attitude error of only 16.7° at the aerobatic point.
We believe that one important factor contributing to this strong performance is the ultralow latency of our system, which enables highly reactive responses to rapidly changing environments.
\begin{table}[t]
	\caption{\textbf{Benchmark Results.}
		In columns TK.P and TK.A, the numbers outside the parentheses denote the mean tracking errors, whereas the \textbf{\textcolor{red}{red}} numbers inside the parentheses represent the maximum tracking errors.
	}
	
	\label{tab:benchmarks}
	\renewcommand{\arraystretch}{1.25}
	\centering
	\scalebox{1.1}
	{
		\begin{tabular}{c|ccccc}
			\hline
			Method   & \begin{tabular}[c]{@{}c@{}}F.T \\ (s)\end{tabular} & \begin{tabular}[c]{@{}c@{}}Aer.P \\ (m)\end{tabular} & \multicolumn{1}{c}{\begin{tabular}[c]{@{}c@{}}Aer.A \\ (deg)\end{tabular}} & \multicolumn{1}{c}{\begin{tabular}[c]{@{}c@{}}TK.P \\ (m)\end{tabular}} & \multicolumn{1}{c}{\begin{tabular}[c]{@{}c@{}}TK.A \\ (deg)\end{tabular}} \\ \hline
			Baseline~\cite{wmy} &16.9 &0.51 &23.4 & 
			\begin{tabular}[c]{@{}c@{}}0.17 \\ (\textbf{\textcolor{red}{0.95}})\end{tabular}
			&
			\begin{tabular}[c]{@{}c@{}}9.80 \\ 
				(\textbf{\textcolor{red}{72.1}})
			\end{tabular}
			\\ \hline
			Proposed & \textbf{12.3}& \textbf{0.35} & \textbf{10.2} &-&-\\ \hline
		\end{tabular}
	}
\end{table}

\begin{table}[t]
	\caption{\textbf{Model Performance in Dynamic Scenes.} 
		Suc.R is the success rate. M.V is the velocity of the moving aerobatic point.
	}	
	\label{tab:dynamic}
	\renewcommand{\arraystretch}{1.25}
	\centering
	\scalebox{1.1}
	{
		\begin{tabular}{c|ccccc}
			\hline
			Cases                                                & \multicolumn{1}{c}{\begin{tabular}[c]{@{}c@{}}M.V=\\ 1.0m/s\end{tabular}} & \begin{tabular}[c]{@{}c@{}}M.V=\\ 1.6m/s\end{tabular} 
			& \multicolumn{1}{c}{\begin{tabular}[c]{@{}c@{}}M.V=\\ 2.2m/s\end{tabular}} & \multicolumn{1}{c}{\begin{tabular}[c]{@{}c@{}}M.V=\\ 2.8m/s\end{tabular}} & \multicolumn{1}{c}{\begin{tabular}[c]{@{}c@{}}M.V=\\ 3.4m/s\end{tabular}} \\ \hline
			Suc.R (\%) &100& 98& 96 &92 &78       \\ \hline
			Aer.P (m)  &   
			0.39	& 0.44 & 0.49&0.61 & 0.63\\ \hline
			Aer.A (\%)
			& 11.5  &12.4 &13.2 &14.5 &16.7 \\ \hline
		\end{tabular}
	}
\vspace{-1.5cm}
\end{table}
\subsection{Real-World Experiments}
\begin{figure}[t]
	\centering
	\includegraphics[width=1\linewidth]{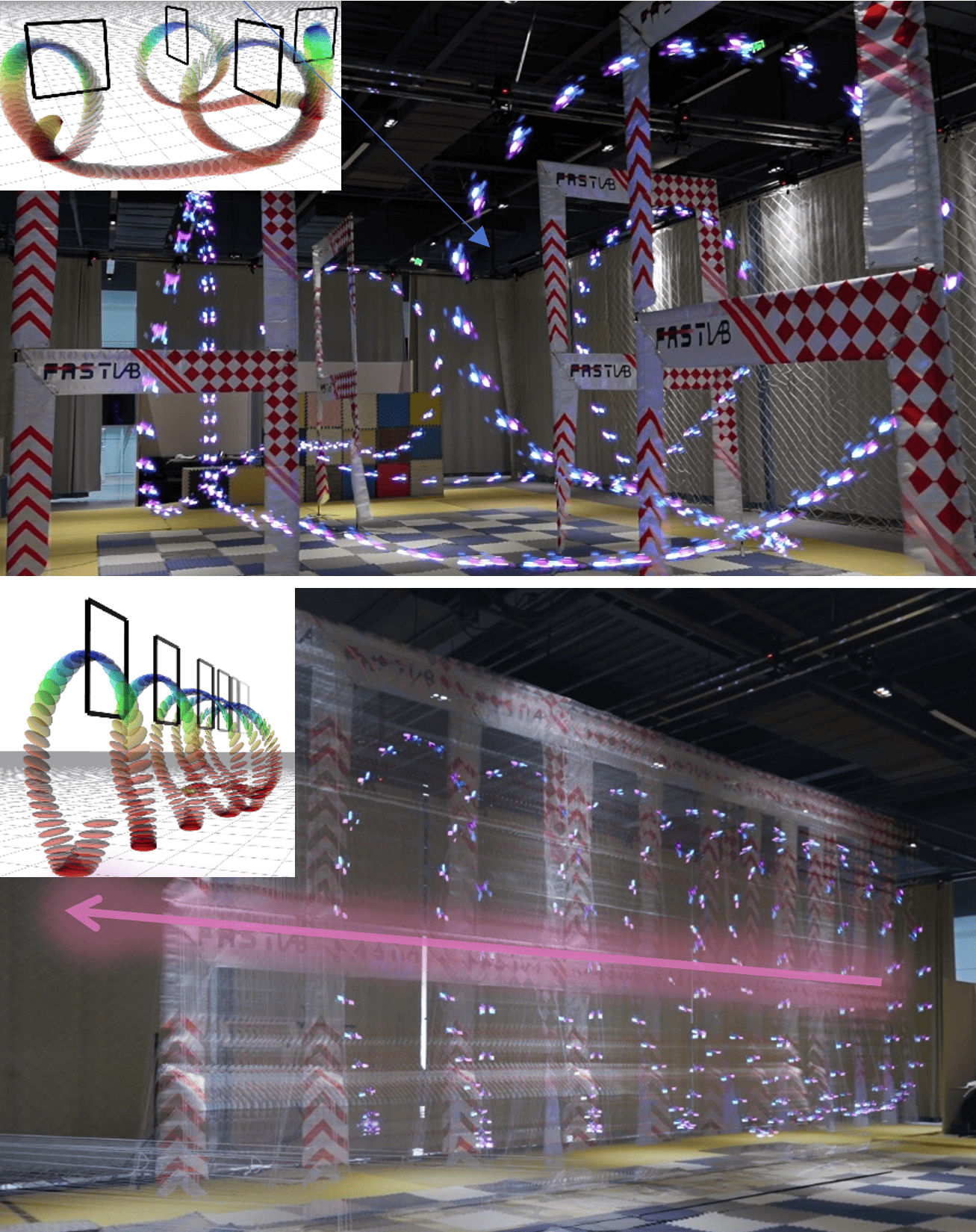}
	\captionsetup{font={small}}
	\caption{
Real-world experiments. The top row shows the static scenario, while the bottom row illustrates the dynamic scenario. Purple arrows indicate the direction of motion.
	}
	\vspace{-0.5cm}
	\label{fig:world}
\end{figure}

In addition to the simulations, we further validate our method in real-world experiments under both static and dynamic scenarios, demonstrating the practical applicability of our approach. To achieve accurate and high-frequency UAV state estimation, we employ an Extended Kalman Filter (EKF) to fuse observations from a  Motion Capture System (MCS) with inertial measurement unit (IMU) data. 
To better demonstrate the experimental results, we construct square-shaped gates to approximately highlight the spatial positions of the aerobatic waypoints. These gates are accurately tracked in real-time by the MCS, from which the precise aerobatic waypoints  are calculated and subsequently provided to the policy network as input.
During these experiments, the network outputs are constrained within predefined practical ranges: with the mass-normalized collective thrust bounded between 0 m/s\textsuperscript{2} and 20 m/s\textsuperscript{2}, roll and pitch angular velocities limited to a range of ±5 rad/s, and yaw angular velocity set within ±3.14 rad/s.
Experimental results from the static scenario (illustrated in the top part of Fig.~\ref{fig:world}) demonstrate that our learned policy can successfully control the UAV through four consecutive gates rapidly, even performing continuous inverted flight. Additionally, this aerobatic maneuver can be repeated seamlessly, demonstrating the robustness and consistency of our method.
In the dynamic scenario experiment, we set a gate translating in its plane at approximately 1.0 m/s and require the UAV to consecutively pass through the moving gate while maintaining an inverted flight orientation, performing challenging power-loop aerobatic maneuvers. Such dynamic and agile aerobatic tasks impose stringent requirements on the responsiveness and accuracy of UAV control systems, as even minor deviations may cause catastrophic crashes. Nevertheless, as illustrated in the bottom part of Fig.~\ref{fig:world}, our method demonstrates fast, reactive, and precise flight control, successfully accomplishing this challenging aerobatic maneuver with notable robustness and reliability.
  A more intuitive experimental demonstration is provided in the supplementary video.
\section{Conclusion}
In this paper, we propose a reinforcement learning-based framework that enables quadrotors to autonomously achieve extreme aerobatic maneuvers.
Through automated curriculum learning and domain randomization, we achieve efficient training and robust zero-shot sim-to-real transfer, validated by real-world experiments demonstrating unprecedented agility. 
In the future, we will utilize generative AI to autonomously generate aerobatic intention without the need for human pre-specification.
\bibliography{ref}

\end{document}